\documentclass{article}

\usepackage[final]{nips_2019}

\usepackage[utf8]{inputenc} 
\usepackage[T1]{fontenc}    
\usepackage{hyperref}       
\usepackage{url}            
\usepackage{booktabs}       
\usepackage{amsfonts}       
\usepackage{nicefrac}       
\usepackage{microtype}      
\usepackage{multirow}
\usepackage{color}
\usepackage{amsmath,amssymb} 
\usepackage{pgfplots}
\pgfplotsset{compat=newest}
\usepackage{pdflscape}
\usepackage{graphicx}
\usepackage[export]{adjustbox}
\usepackage{algorithm}
\usepackage{mathtools}
\usepackage{algorithmic}
\usepackage{wrapfig,booktabs}

\usepackage{array}

\usepackage{wrapfig}

\title{ExpertMatcher: Automating ML Model Selection for Users in Resource Constrained Countries}

\author{
  Vivek Sharma \\
  MIT\\
  \texttt{vvsharma@mit.edu} \\
  \And
  Praneeth Vepakomma \\
  MIT \\
  \texttt{vepakom@mit.edu} \\
  \And
  Tristan Swedish \\
  MIT \\
  \texttt{tswedish@mit.edu} \\
  \AND
  Ken Chang \\
  MIT \\
  \texttt{kenchang@mit.edu} \\
  \And
  Jayashree Kalpathy-Cramer \\
  MGH/Harvard Medical School \\
  \texttt{kalpathy@nmr.mgh.harvard.edu} \\
  \And 
  Ramesh Raskar \\
  MIT \\
  \texttt{raskar@mit.edu} \\
}

\begin{document}

\maketitle
\begin{abstract}
In this work we introduce ExpertMatcher, a method for automating deep learning model selection using autoencoders. Specifically, we are interested in performing inference on data sources that are distributed across many clients using pretrained expert ML networks on a centralized server. The ExpertMatcher assigns the most relevant model(s) in the central server given the client's data representation. This allows resource-constrained clients in developing countries to utilize the most relevant ML models for their given task without having to evaluate the performance of each ML model. The method is generic and can be beneficial in any setup where there are local clients and numerous centralized expert ML models.
\end{abstract}

\section{Introduction}

In developing countries, there are often scenarios where (1) the amount of data is scarce and training a deep model from scratch is not feasible; (2) there is a lack of computational resources to train effective deep learning models; or (3) there are beginner users who have minimal or no expertise in ML. In contrast, the abundance of pre-trained models makes inference accessible to users who do not have access to data, computational resources, or domain expertise. Thanks to modern cloud infrastructure and distributed learning methods~\citep{vepakomma2018split,vepakomma2019reducing,vepakomma2018no,chang2018distributed}, ``expert model hubs'' can make inference available to any remote client connected to the internet. This allows the remote client to utilize powerful expert models for their local applications.

Model selection is an important problem that can not be simply sidestepped by training a large capacity multi-task model to perform the inference directly. In practice, training a single model for multiple tasks while maintaining performance across tasks is a challenging problem due to catastrophic forgetting, a phenomenon in which sequential learning of a new task leads to lower performance on previously learned tasks~\citep{goodfellow2013empirical}. As such, most deep learning models are trained to be domain-specific. Even within a domain, there may be differences in the performance of models trained with different datasets due to variations in demographics, class prevalence, data collection instrument, and data acquisition settings~\citep{zech2018variable,tomavsev2019clinically,albadawy2018deep,ting2017development}.

The key idea in this paper is to match input data based on its likelihood of being drawn from the distribution of the expert model training data. To this end, we seek a general representation of our dataset that can be easily compared to an input data sample. We propose an autoencoder~(AE) based expert matcher that learns the underlying representation for a given dataset and automatically triggers the expert network when a clients data representation matches the AE representation. This allows the client to effectively utilize centralized expert networks for the given task. Figure~\ref{fig:pipeline} sketches an overview of the proposed method. Our main contribution of the paper are, (1) we describe the landscape of expert matching; and (2) we propose an expert matcher for automatic ML model selection for users in resource constrained settings when sharing client data with the server is not a concern. Our proposed approach is evaluated on 6 benchmark datasets: STl-10~\citep{stl}, MNIST~\citep{mnist}, HAR~\citep{har}, Reuters~\citep{reuters}, Non Line of Sight~\citep{nlos} and Diabetic Retinopathy~\citep{db}. We show that ExpertMatcher can be used to match both the task (coarse-Level) and class (fine-grained). Matching the class mimics the scenario where you have multiple trained models for the same task, each with a deferentially biased training set. Fine-grained matching allows for finding of the model trained on data with a class distribution most similar to the local data.

The rest of the paper is organized as follows. In Section~\ref{sec:relwork}, we discuss related work. Section~\ref{sec:method} describes our
proposed approach. Experimental results and their analysis
are presented in Section~\ref{sec:experiments}. 

\vspace{-3mm}
\section{Related Work} \label{sec:relwork}
Jacobs et al.~\citep{jacobs} proposed the first examples of using multiple expert models, each expert model handling a subset of tasks. They trained an adaptive mixture of experts for speaker vowel recognition and used a gating network to determine which of the networks should be used for each sample. The gating function learned what training sample needed to be passed to which expert. For a more detailed review, please see~\cite{jacobs1995methods,jacobs1997bias}. In order to avoid the gating function, in~\citep{hinton,ahmed} trained a mixture of one oracle model that provides common knowledge to many specialist experts in the form of shared features. The oracle model acts as a gating function for passing the sample to the expert network. The oracle model sees the whole training data to do an accurate assignment and also needs to be retrained when a new dataset is added. Similar is the case with the~\citep{aljundi}, where the author also learns a gating function to make expert network assignments. We are inspired by all of these works. In contrast to these prior works, our work differs substantially in scope and technical approach. We use a simple autoencoder and do the expert network assignment based on the reconstruction loss for each sample. 

\begin{wrapfigure}{r}{0.5\textwidth}
\vspace{-8mm}
{\includegraphics[width=0.5\columnwidth]{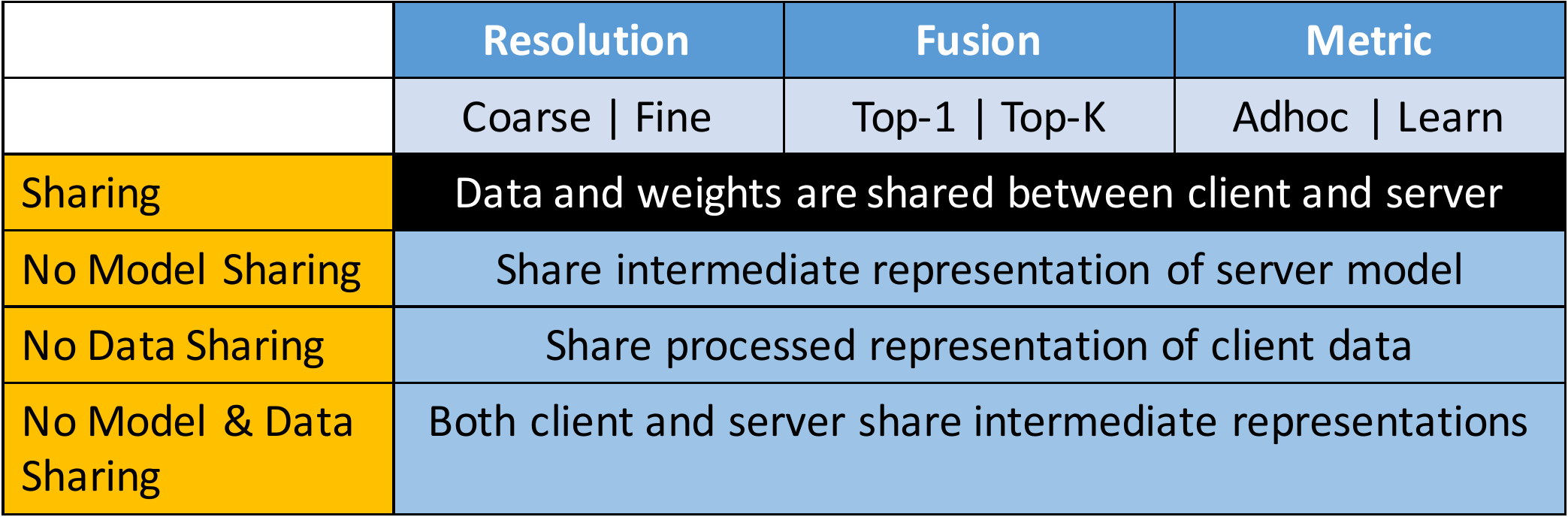}} 
\vspace{-5mm}
\caption{Landscape of ExpertMatcher}
\label{fig:landscape}
\vspace{-5mm}
\end{wrapfigure}

\vspace{-3mm}
\section{ExpertMatcher Problem} \label{sec:method}

We aim to assign an expert network to a given sample from the client data with the goal to dynamically handle sample-specific correction.  

Figure~\ref{fig:landscape} shows the landscape of the ExpertMatching problem. Given the distributed nature of the problem, it is important to consider what data/model the client and server share with each other and the respective privacy trade-offs. In general, there are three main aspects that guides the ExpertMatcher, (1) \textbf{Resolution}: coarse and fine level assignment; (2) \textbf{Fusion}: using top-1 or top-K expert models; (3) \textbf{Metric}: adhoc (e.g. MSE or cosine similarity) or learnable assignment metric. In this work we consider the first sharing scenario (row one in Figure~\ref{fig:landscape}), where the client and server share the data and model.

\begin{wrapfigure}{r}{0.5\textwidth}
\centering
\vspace{-9mm}
{\includegraphics[width=0.4\columnwidth]{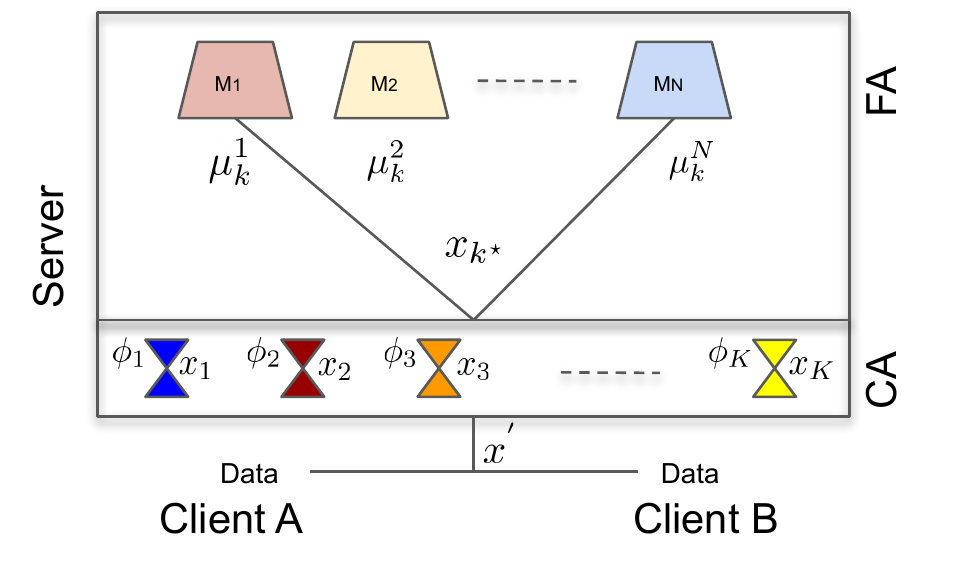}} 
\vspace{-5mm}
\caption{\textbf{ExpertMatcher}. Overview of the proposed unified distributed learning using expert matcher. The ExpertMatcher works hierarchically, where it first triggers the best model for the clients data representation in  coarse level assignment~(CA), and then followed by fine-level handing the clients data  in fine-grained assignment. }
\label{fig:pipeline}
\vspace{-10mm}
\end{wrapfigure}


Our approach has three key qualities: (i) \textbf{modularity}:  the client can easily benefit when new expert models are added on the server; (ii) \textbf{efficiency}: it does not need the client to train a model; and (iii) \textbf{expert-free}: specialized or expert knowledge can help the clients without any overhead to solve their task at hand.

We describe our proposed solution to the ExpertMatcher problem as follows, considering coarse and fine matching resolution:  (1)  Coarse-level expert matcher~(CA), and (2) Fine-grained expert matcher~(FA). Figure~\ref{fig:pipeline} sketches the pipeline.

\textbf{Coarse-level expert matcher~(CA).} We assume, we have $K$ pre-trained autoencoders~(AE) $\phi_{K}$ for $K$ datasets on the server. The AEs are trained using reconstruction loss i.e. using mean squared error~(MSE) loss between the target data $x^{'}$ and the network's predicted output data $\hat{x}$, $\phi_{k}: x^{'} \rightarrow \hat{x_{k}}, k \in \{1,\ldots,K\}$. The reconstruction loss is given by $loss_{k} = MSE(x^{'}, \hat{x_{k}})$. The intermediate features extracted from a hidden layer for a sample $x^{'}$ is given as $x_{k}=\phi_{k}(x^{'})$.  To compare the reconstruction output $\hat{x_k}$ with the ideal $x^{'}$, we use MSE loss as a measure of quality, although we note that more complex loss functions could be used. 

For expert assignment of client data, we assign an AE, $k^{\ast} \in \{1,\ldots,K\}$ which has minimum reconstruction error, that means the semantic feature space of client data matches to that of the underlying AE space.

\textbf{Fine-grained expert matcher~(FA).} Given that an autoencoder is trained on a dataset with $N$ classes e.g. MNIST we have 10 letters ($N=10$). Once an AE is trained for each dataset, we obtain a hidden representation of each class and compute an average representation of each class in the dataset e.g. for MNIST we have 10 mean representations one for each class $\mu^{n} \in \mathbb{R}^{d} , n \in \{1,\ldots,N\}$, where $d$ is hidden layer dimension, and $N$ is the number of object classes that varies for each dataset.

For fine level expert assignment of client data $x^{'}$, we assign an expert model $M_{n}, n \in \{1,\ldots,N\}$ which has the highest cosine similarity of $x_{k^{\ast}}$ with $\mu^{n}$.

This mimics a scenario in which there are different centralized models that have biased training sets (ie class imbalances) and you want to locate the model trained on data that has the most similar class distribution to your local dataset.

In the current setup, privacy is not a concern. The client shares their dataset with online services.

\vspace{-3mm}
\section{Experiments} \label{sec:experiments}
Our experimental setting is illustrated in Figure~\ref{fig:pipeline}: a random sample is selected by the client from a collection of data sources, and the ExpertMatcher ``Server'' must select correct corresponding model.  We demonstrate that the ExpertMatcher can distuinguish between a set of data sources which cover domains such as text, digits, objects, sensor, and medical images. We first introduce the datasets and evaluation metric, followed by a thorough comparison of the proposed method for both coarse assignment~(CA) and fine-grained assignment~(FA). 

\begin{wrapfigure}{r}{0.5\textwidth}
\vspace{-8mm}
{\includegraphics[width=0.5\columnwidth]{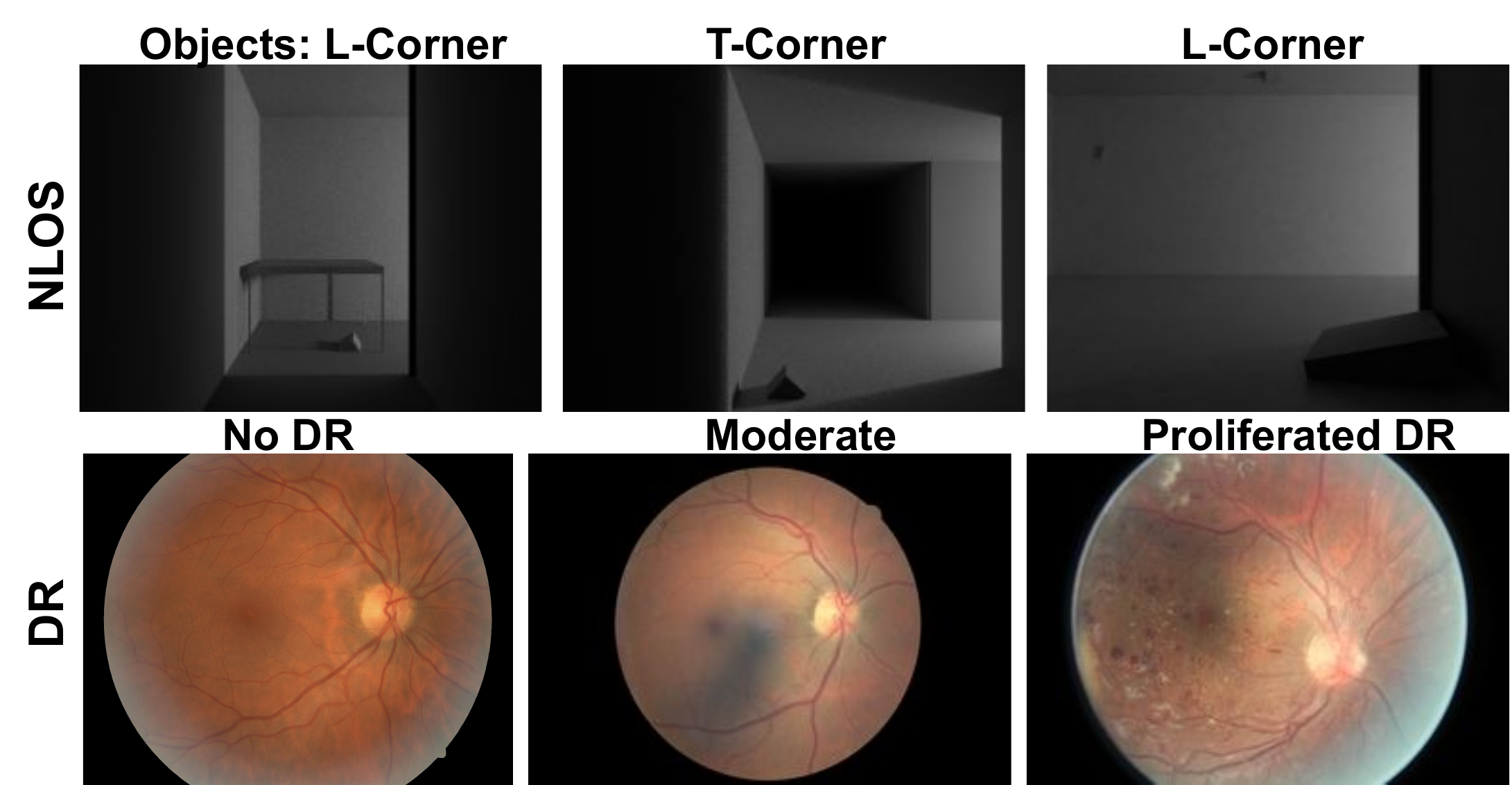}} 
\vspace{-5mm}
\caption{Example images for a few samples from our Non-Line of Sight~(NLOS) and Diabetic Retinopathy~(DB) datasets. We show one sample per class for both datasets. The coarse similarity between classes makes the datasets challenging.}
\label{fig:dataset}
\vspace{-8mm}
\end{wrapfigure}

\textbf{Datasets.} 
We present a summary of the dataset used in this work in Table~\ref{table:stats}.  We report the number of samples used by the server to train an autoencoder for each dataset. We also report the number of samples for both clients. We divide the dataset into non-overlapping splits 50/25/25\% for Server, Client A, and Client B correspondingly. Additionally, some datasets have varying class distribution, indicated by the cluster skew between the largest class (LC)  to the smallest class (SC).  Figure~\ref{fig:dataset} shows a few examples Non-Line of Sight~(NLOS) and Diabetic Retinopathy~(DB) samples included in our dataset.

\begin{table*}[ht]
\small
\tabcolsep=1.5mm
\begin{center}
\caption{\textbf{Datasets}. ``\#S'' denotes the number of samples, ``\#C'' denotes the true number classes/clusters, and largest class (LC)
/ smallest class (SC) is the class balance percent of the given data.} 
\label{table:stats}
\resizebox{10cm}{!}{
\begin{tabular}{l|cccccc|c}
\toprule
&	 STL-10 & MNIST & HAR &  REUTERS & NLOS & DB& ALL  \\
\cmidrule{1-7}
Type & Object & Digits & Sensors & Text & Sensor & Biological & \\
\#C & 10 & 10 & 6 & 4 & 3 & 3 & \\
\#S & 13k &10k & 10299 &10k &45096  & 3540 &\\
Dim. & 32px & 28px & 561 & 2000 & $640\times480$ & 512px \\
LC/SC~(\%) & 10/10 &  11.35/8.92 & 19/14 & 43.12/8.14 &  33.33/33.33& 33.33/33.33&\\
\midrule
Server  &   6500&   5000&   5151&   5000& 22548 & 1770    \\ 
Client A &   3250&   2500&   2574&   2500& 11274 & 885 & 22983\\
Client B &   3250&   2500&   2574&   2500& 11274 & 885 & 22983\\
\bottomrule
\end{tabular}}
\end{center}
\vspace{-5mm}
\end{table*}

\textbf{Evaluation Metric.} For CA, we use minimum reconstruction error (i.e. MSE) for making a model assignment and then computing the accuracy between predicted dataset and target dataset. For FA, we use maximum cosine similarity for making a class assignment and then computing the accuracy between predicted class and target class.

\textbf{Multiple Clients} The performance is evaluated for two subsets of non-overlapping client data, providing a rough measure of the expected variance of ExpertMatcher performance if clients draw different samples from the same underlying distribution.

\textbf{Implementation Details.} We adopt the input images by resizing to $28 \times 28$, then flattening it to 784 dimensions. For HAR and Reuter, we apply 1D adaptive average pooling (\texttt{AdaptiveAvgPool1d}) to transform the input data to 784 dimensions. \textbf{AE:} Our AE uses a single-layer MLP encoder-decoder ($\mathbb{R}^{784} \rightarrow \mathbb{R}^{128} \rightarrow \mathbb{R}^{784}$. \textbf{MLP:} The network comprises of fully-connected layers ($\mathbb{R}^{784} \rightarrow \mathbb{R}^{256} \rightarrow \mathbb{R}^{128} \rightarrow \mathbb{R}^{C}$) using $C$-way softmax layer, where $C$ is the number of dataset categories. The AE and MLP model parameters are trained using Adam optimizer. We initialize the learning rate with $10^{-2}$ and manually decrease by a factor of 10 every 15 epochs. The maximum number of epochs is set to 45. We use batch normalization. 
\vspace{-0.3cm}

\subsection{Coarse-level dataset assignment~(CA)}
\begin{wraptable}{r}{5.5cm}
\small
\tabcolsep=1mm
\begin{center}
\vspace{-13mm}
\caption{ExpertMatcher. Average accuracy~(\%) for assigning STL-10, MNIST, HAR, REUTERS datasets.} 
\label{table:ae_mlp}
\vspace{-3mm}
\begin{tabular}{l|cc|c}
\toprule
Method & Client A & Client B & Samples \\
\toprule
AE-MSE           & 99.94 & 99.91 &10824\\
MLP-Softmax         & 99.95 & 99.97&10824\\
\bottomrule
\end{tabular}
\end{center}
\vspace{-6mm}
\end{wraptable}

In Table~\ref{table:ca}, we show the results for the coarse level dataset assignment. We can observe that autoencoders are very effective in learning the semantics of the underlying data representation even with relatively low input resolution (e.g. 28 $\times$ 28 for images). We can notice that both Client A and Client B are 99\% accurately assigned to their corresponding Autoencoders.

In Table~\ref{table:ae_mlp}, we compare results of autoencoder using MSE loss based assignment to MLP using softmax classification for dataset assignment. It is interesting to observe that AE based assignment is equally good as MLP assignment. However, note that with MLP it is not possible to do fine-grained assignment until and unless we have learned MLP to do both dataset and class recognition as a multi-class classification problem.
\vspace{-3mm}
\begin{table}[ht]
  \caption{Coarse level dataset assignment using MSE loss as the assignment metric for computing accuracy~(\%).}
  \label{table:ca}
  \centering
  \vspace{-6mm}
  \resizebox{10cm}{!}{
  \begin{tabular}{l|cccccc|c}
    \toprule
         & MNIST &  STL-10  & HAR & REUTERS & NLOS & DB & Average   \\
    \midrule
    Client A     & 100.0 & 100.0  & 100.0  & 99.64  & 99.92  & 96.49 & 99.34 \\
    Client B     & 100.0 & 100.0  & 100.0  & 99.56 & 99.89  & 95.36 & 99.13    \\
    \bottomrule
  \end{tabular}}
  \vspace{-0.6cm}
\end{table} 

\begin{wraptable}{r}{5.5cm}
\small
\tabcolsep=1mm
\begin{center}
\vspace{-13mm}
  \caption{Fine-grained class assignment using cosine-similarity as the assignment metric for computing accuracy~(\%).}
  \label{table:fa}
  \begin{tabular}{lccc}
    \toprule
    Dataset     & \#Classes & Client A     & ClientB \\
    \midrule
    MNIST   &10 &  84.36 & 83.40  \\
    NLOS    &3  &  71.78 & 71.26 \\
    DB      &3  &  41.47 & 44.41 \\
    \bottomrule
\end{tabular}
\end{center}
\vspace{-10mm}
\end{wraptable} 

\subsection{Fine-grained class assignment~(FA)}
In Table~\ref{table:fa}, we show the results for the fine-grained class assignment. In Table~\ref{table:fa}, we observed that autoencoder seem to learn the underlying dataset representation, here in Table~\ref{table:fa} we observe that autoencoders are effective at learning the class identity representation too. Note that fine-grained recognition is a very hard problem, in Figure~\ref{fig:dataset} some sample examples are shown, which primarily shows that the class categories are very similar.
\vspace{-3mm}

\section{Conclusion}

We demonstrate ExpertMatcher, a model selection method for deep learning models in a distributed setting. Our method matches data inputs from remote clients to pre-trained models on a central server for inference. ExpertMatcher trains autoencoders for each model's training dataset, and uses either reconstruction error or cosine similarity of the autoencoder's bottleneck layer for either coarse or fine-grained matching respectively. Our method is able to distinguish between many common benchmark datasets at a coarse and fine level with reasonable accuracy. We also demonstrate that our method can be applied to practical domain specific models for Non-line-of-sight and Diabetic Retinopathy.

\textbf{Acknowledgements}
V. Sharma would like to thank Karlsruhe House of Young Scientists~(KHYS) for funding his research stay at MIT.







\bibliographystyle{authordate1} 
\bibliography{main}

\end{document}